\documentclass[sigconf]{acmart}
\AtBeginDocument{%
  \providecommand\BibTeX{{%
    \normalfont B\kern-0.5em{\scshape i\kern-0.25em b}\kern-0.8em\TeX}}}

\copyrightyear{2021}
\acmYear{2021}
\setcopyright{iw3c2w3}
\acmConference[WWW '21]{Proceedings of the Web Conference 2021}{April 19--23, 2021}{Ljubljana, Slovenia}
\acmBooktitle{Proceedings of the Web Conference 2021 (WWW '21), April 19--23, 2021, Ljubljana, Slovenia}
\acmPrice{}
\acmDOI{10.1145/3442381.3450020}
\acmISBN{978-1-4503-8312-7/21/04}

\acmSubmissionID{fp5183601}

\usepackage{multirow}
\usepackage{color}
\usepackage{amsmath,amsfonts,bm}

\newcommand{\model}{\textsc{DeePRed}}
\newcommand{\lstm}{\textsc{lstm}}
\newcommand{\timelstm}{\textsc{TimeLstm}}
\newcommand{\latentcross}{\textsc{LatentCross}}
\newcommand{\ctdne}{\textsc{ctdne}}
\newcommand{\rrn}{\textsc{rrn}}
\newcommand{\deepcoevolve}{\textsc{DeepCoEvolve}}
\newcommand{\jodie}{\textsc{Jodie}}
\newcommand{\jodienf}{\textsc{JodieNF}}

\newcommand{\splitter}{\textsc{Splitter}}
\newcommand{\deepwalk}{\textsc{Deepwalk}}
\newcommand{\nodetovec}{\textsc{Node2vec}}
\newcommand{\lne}{\textsc{line}}

\def\vb{{\bm{b}}}

\def\ve{{\bm{e}}}

\def\vh{{\bm{h}}}
\def\vi{{\bm{i}}}

\def\vn{{\bm{n}}}

\def\vr{{\bm{r}}}

\def\vu{{\bm{u}}}
\def\vv{{\bm{v}}}

\def\vz{{\bm{z}}}

\def\mA{{\bm{A}}}

\def\mE{{\bm{E}}}
\def\mF{{\bm{F}}}

\def\mI{{\bm{I}}}

\def\mS{{\bm{S}}}

\def\mW{{\bm{W}}}
\def\mX{{\bm{X}}}

\def\gL{{\mathcal{L}}}
\def\gS{{\mathcal{S}}}

\def\sI{{\mathbb{I}}}

\def\sL{{\mathbb{L}}}

\def\sR{{\mathbb{R}}}

\def\sU{{\mathbb{U}}}

\setcopyright{iw3c2w3}

\begin{document}

\title{Dynamic Embeddings for Interaction Prediction}

\author{Zekarias T. Kefato}
\email{zekarias@kth.se}
\affiliation{%
  \institution{KTH Royal Institute of Technology}
  \city{Stockholm}
  \state{Sweden}
}

\author{Sarunas Girdzijauskas}
\email{sarunasg@kth.se}
\affiliation{%
  \institution{KTH Royal Institute of Technology}
  \city{Stockholm}
  \state{Sweden}
}

\author{Nasrullah Sheikh}
\email{nasrullah.sheikh@ibm.com }
\affiliation{%
  \institution{IBM Research – Almaden}
  \city{San Jose}
  \state{USA}
}

\author{Alberto Montresor}
\email{alberto.montresor@unitn.it}
\affiliation{%
  \institution{University of Trento}
  \city{Trento}
  \state{Italy}
}

\renewcommand{\shortauthors}{Kefato and Girdzijauskas, et al.}

\begin{abstract}
  In recommender systems (RSs), predicting the next item that a user interacts with is critical for user retention.
  While the last decade has seen an explosion of RSs aimed at identifying relevant items that match user preferences, there is still a range of aspects that could be considered to further improve their performance.
  For example, often RSs are centered around the user, who is modeled using her recent sequence of activities. 
  Recent studies, however, have shown the effectiveness of modeling the  \emph{mutual} interactions between users and items using separate user and item embeddings.  
  
  Building on the success of these studies, we propose a novel method called {\model} that addresses some of their limitations. 
  In particular, we avoid recursive and costly interactions between consecutive short-term embeddings by using long-term (stationary) embeddings as a proxy.
    This enable us to train {\model} using simple mini-batches without the overhead of specialized mini-batches proposed in previous studies. 
    Moreover, \model's effectiveness comes from the aforementioned design and a multi-way attention mechanism that inspects user-item compatibility.
  Experiments show that {\model} outperforms the best state-of-the-art approach by at least 14\% of Mean Reciprocal Rank (MRR) on next item prediction task, while gaining more than an order of magnitude speedup over the best performing baselines.
  Although this study is mainly concerned with temporal interaction networks, we also show the power and flexibility of {\model} by adapting it to the case of static interaction networks, substituting the short- and long-term aspects with local and global ones.

\end{abstract}




\maketitle

\section{Introduction}
\label{sec:intro}
Vital to the success of a number of real-world recommender systems (RS) is the ability to predict future interactions between entities based on their previous interaction history.
In many recommender systems, effective user-item interaction prediction enables end-users to sift through an overwhelming number of choices.
In addition, in biology, pharmacology and related fields, interaction prediction between biological and chemical compounds has been explored to better understand unknown bio-chemical interactions~\cite{BUZA2017284,YOU201990,10.1093/bioinformatics/bty294}.

In this paper, we are primarily interested in temporal interaction networks between two sets of entities (\emph{users} and \emph{items}).
The terms cover a variety of notions, \emph{e.g.} users could be customers in an e-commerce system, or accounts on Reddit, YouTube or Spotify; items could be products, posts, media produced or consumed by users.

Given a set of observed interactions between users and items, predicting possible future interactions is an increasingly important and challenging task.
The goal of this paper is to introduce a new method to predict the next items that users interact with, based on their previous history of interaction.
We model our problem through bipartite temporal interaction networks, as they can naturally and effectively represent user-item interactions over time.


\paragraph{\textbf{Existing studies}} In the context of RS, several approaches have been proposed to predict future items a user is likely to interact with, providing encouraging results~\cite{10.1145/3018661.3018689,Wu_2019,hidasi2015sessionbased,ijcai2019-547,10.1145/3397271.3401142,10.1145/2988450.2988452,kumar2019predicting,10.1145/2959100.2959190,dai2016deep}.
Often times, however, the focus is on modeling users, while the user-item interaction dynamics that provide a richer signal is overlooked~\cite{10.1145/3331184.3331267}.
In several cases, RNNs and other models suitable for sequences were used to train a predictive model over the item sequence corpus.

Recently, studies have shown how to mutually model both user and items based on bipartite interaction networks and demonstrate significant improvement over existing methods~\cite{kumar2019predicting,dai2016deep}.
Unlike previous approaches, they have employed mutually recursive RNNs that are more capable to model the user-item interaction dynamics.
While they use two types of embeddings, long-term and short-term, the former is just a fixed one-hot vector and the latter is the real core of their models, that it is used to capture recent user preferences and item properties.
Moreover, these approaches work by recursively computing the short-term embedding at time $t$ based on the embedding at time $t-1$ , which leads to sequential training that proved to be a bottleneck as the network scales up. Even if recent work has introduced a mini-batch training algorithm, the overhead is not completely alleviated yet~\cite{kumar2019predicting}.


\paragraph{\textbf{This study}} We propose a novel algorithm called \model~\footnote{The source code is available at \href{https://github.com/zekarias-tilahun/deepred}{https://github.com/zekarias-tilahun/deepred}} (\underline{D}ynamic  \underline{E}mbeddings  for Int\underline{e}raction \underline{Pred}iction). 
{\model} provides a simple yet powerful way of modeling short-term interaction behaviours that removes the aforementioned recursive dependency for efficient training.
This is achieved by decoupling the \emph{learnable} user or item embeddings into long-term and short-term embeddings, 
in order to capture both stationary and transitory interaction patterns.
Furthermore, {\model} computes separate embeddings from the point of view of both: users and items.
Henceforth, although our discussion mostly covers users, the same principles can be applied to items unless explicitly stated otherwise.

\begin{figure}[t!]
    \centering
    \includegraphics[scale=0.4]{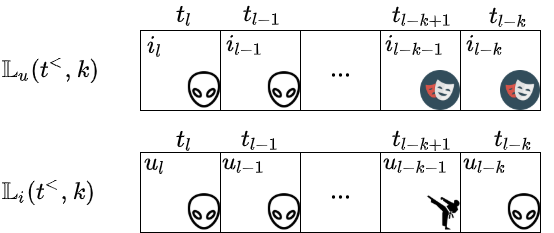}
    \caption{An illustration of participants and the context of the participants of each event for the $k$ recent interaction events of user $u$ ($\sL_u(t^<, k)$)  and an item $i$ ($\sL_i(t^<, k)$). E.g. the last interaction of $u$ was with item $i_l$ from SciFi context (alien icon) and $i$ has interacted with user $u_l$ from SciFi context.}
    \label{fig:history_illustration}
\end{figure}

The key idea behind the effectiveness of {\model} is that, each time a user interacts with an item, it is modeled using a sequence of $k$ recent items she interacted with, which reflects a context of interaction.
For example, Fig.~\ref{fig:history_illustration} shows the context of the $k$ recent interaction events of a user $u$ and an item $i$.
We see that the two most recent interactions at $t_l$ and $t_{l-1}$ are within the context of SciFi, for both $u$ and $i$.
That is, the last two items that $u$ has interacted with are relevant to the theme of SciFi.
Thus, the long-term (contextually stationary) embedding of the items (e.g. $i_l=Spider Man$ and ${i_{l-1}=Terminator}$) at $t_l$ and $t_{l-1}$ are used to encode such context.

Similar to previous work~\cite{kumar2019predicting,dai2016deep}, we use two mutual RNNs that capture the interaction and temporal patterns within a history sequence (the $k$ most recent interaction events), and generate high-level features.
However, unlike previous work, the two RNNs share the same model parameters and are not recursively dependent.

Finally, the power of {\model} comes from a multi-way attention mechanism that we employ to capture the user-item interaction signal, to check whether the short-term history ($k$ most recent interactions) of a user and an item are compatible using attention weights.
The weights are then used as feature selectors over the high-level features and predict the short-term embeddings.
In \model, each interaction produces a new instance of short-term embedding for both the user and item.
This gives {\model} the power to reason based on consistent behaviours as opposed to rare events, and it is in contrast to~\cite{kumar2019predicting} that updates the existing ones. 
Besides its qualitative power, predicting short-term embeddings as opposed to interaction probabilities is another choice in our design that boosts \model's efficiency.

The last but not the least aspect of {\model} is that it can be seamlessly extended to tackle static interaction networks.
This is achieved by replacing long and short-term aspects with global and local ones, based on a sample of interactions as opposed to the latest (recent) ones.

Our contributions are the following:
\begin{itemize}
    \item \textbf{Novelty}: We propose a novel algorithm that captures user (item) preferences over time by modeling users (items) using their recent interaction history. 
    By leveraging the decoupling of the learnable embeddings, we employ \emph{non-recursive} mutual RNNs to capture interaction and temporal patterns within the histories.
    Furthermore, an attention mechanism is used to inspect user-item compatibility allowing to significantly improve the predictive performance of our approach.
    \item \textbf{Empirical results:} With respect to the state of the art, our results show at least a 14\% gain on mean reciprocal rank, measured on three real-world and publicly available datasets.
    \item \textbf{Efficiency}: As a result of eliminating the recursive self-dependency between short-term embeddings at different time steps, {\model} achieves more than one order of magnitude speedup over the best performing SOTA methods. 
    \item \textbf{Easy extension to static networks}: Though the focus of this study is on temporal interaction networks, we have shown that {\model} is seamlessly extendable to static interaction networks using three real-world datasets.
\end{itemize}

\section{Modeling Preliminaries}
\label{sec:modeling_interactions}

The focus of this study is to model temporal interaction networks; yet, our proposal could be adapted to static networks with little effort. We therefore show first the general model, and then we show how to specialize it for the static case.

We take an ordered set $\sL$ containing a log of interactions between a set of users $\sU$  and a set of items $\sI$, where ${L = \vert \sL \vert }$, ${U = \vert \sU \vert}$, and ${I = \vert \sI \vert}$.
An event $e = (u, i, t) \in \sL$ records an interaction between a user $u$ and an item $i$ at time $t$.
Events associated with users and items are intrinsically ordered by time.
Let $\sL_u$ be the set of all interaction events of user $u$, such that  ${\sL_u = \lbrace e_1, e_2, \ldots, e_l \rbrace}$ and events are intrinsically ordered, that is if two events ${e_j = (u, i_j, t_j)}$ and ${e_k = (u, i_k, t_k)}$ are such that $j \leq k$, then $t_j \leq t_k$.

In predicting future interactions between users and items, generally, both long-term and short-term interaction behaviours are commonly used~\cite{10.5555/3172077.3172393,10.1145/3159652.3159727,DBLP:journals/corr/DaiWTS16}.
However, short-term behaviours are mostly favored to have a strong impact on follow-up interactions.
We adopt a similar assumption and model user and item preferences from both long-term and short-term perspectives.
The long-term preferences are captured through the complete interaction histories of users/items.
For a user $u$ and an item $i$, $\sL_u$ and $\sL_i$ denote their complete interaction history, respectively.

Although user preferences are normally considered to change over time~\cite{10.1145/3018661.3018689}, we assume that users usually have a dominant (stationary) preference, which remains unchanged.
However, as their preferences change over time depending on recent actions, users have a tendency to do related actions.
For instance, in movie RS, a particular genre might be preferred by a user at any given time. 
More importantly, however, one is likely to show interest in movies of different genres based on mood, events in her life (e.g. marriage, childbirth, trauma) and seasons (e.g. Christmas, Summer)~\cite{10.1145/3018661.3018689}.
Thus, the most recent watching behaviors have a stronger impact than the old preferences over the next movie that a user is likely to watch.

To capture recent preferences, in line with~\cite{ijcai2019-607,10.1145/3159652.3159727}, we use the $k$ most recent  interaction events.
Unlike some studies~\cite{hidasi2015sessionbased,kang2018selfattentive,Wu_2019,10.1145/2959100.2959190}, however, we assume that the $k$ most recent interaction events from both the user \textit{and} the item influence the next user-item interaction.
Later, we shall discuss the details of the benefit of this design choice. 
Thus, the $k$ most recent interactions of each user ${u \in \sU}$ and item ${i \in \sI}$, respectively, before a given time $t$ are identified by: 
\begin{align*}
    \sL_u(t^<, k) = \lbrace i_j, \Delta_j : (u, i_j, t_j) \in \sL_u, t_j < t, j = l-k, \ldots, l \rbrace \\
    \sL_i(t^<, k) = \lbrace u_j, \Delta_j : (u_j, i, t_j) \in \sL_i, t_j < t, j = l-k, \ldots, l \rbrace
\end{align*}
where $\Delta_j = t-t_j$ captures the hotness (recency) of the $j^{th}$ event. 

For static networks, we simply strip out time from $\sL$; any subsets thereof become unordered. 
In this case, ${\sL_u(k) \subseteq \sL_u}$ and ${\sL_i(k) \subseteq \sL_i}$ simply denote a sampled set of $k$ events from observed events $\sL_u$ and $\sL_i$, respectively.


\paragraph{\textbf{Research question}} The main question of this study is: given an ordered set of observed events $\sL_\mathcal{O}$, can we design an efficient algorithm that effectively predicts future interactions in temporal interaction networks?
In addition, can we make it flexible enough to be applicable to static interaction networks?

\section{\model}
\label{sec:model}

\begin{figure}[t!]
    \centering
    \includegraphics[scale=0.4]{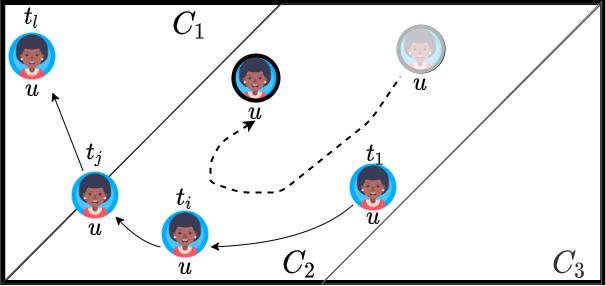}
    \caption{An illustration of the evolution of the short-term and long-term embeddings of user $u$ in a context space, which is further divided into smaller sub-spaces reflecting a context or theme ($C_1, C_2, C_3$). The dotted arrow indicates the trajectory of the long-term embedding (indicated in black circle). The short-term embeddings of the user are annotated with timestamps, which is associated with the interaction that generated them.}
    \label{fig:context_space_and_illustration}
\end{figure}

The proposed algorithm, \model, captures both stationary and transitory preferences of users and items in interaction networks, by maintaining two dynamic embeddings, one long-term and one short-term, in a latent context-space $\gS$. 
The main hypothesis in {\model} is that an underlying hidden context-space $\gS$ is considered to have been generated as a result of interactions between users and items.
This space is assumed to be thematically divided into different regions that are associated to a particular theme or context.
For an intuitive understanding of $\gS$ in \model, let us consider an illustration shown in Fig.~\ref{fig:context_space_and_illustration}. 
To simplify our discussion, suppose $\gS$ is a 2-dimensional euclidean space, which is further divided into three different thematic regions, $C_1, C_2, C_3$.
The notion of a theme/context is related to user interests (preferences) and item properties.

The two dynamic embeddings are updated at every user-item interaction, both for users and items.
Since {\model} applies the same procedure for both, the following discussion is given from a user's perspective.
Suppose user $u$ has interacted with an item relevant to context $C_2$ at time $t_1$.
To reflect such behavior, we start by initializing the user's long-term and short-term embeddings, which are located within the same context $C_2$.

As time progresses, when the user interacts with different items, new instances of the short-term embeddings are generated by keeping the previous ones.
The new instances are shown in the figure along with a timestamp associated to the interaction, which caused the current embedding,  and a solid-line trajectory.
The motivation for keeping the embeddings comes from a need to maintain a smooth short-term embedding that reflects the ``normal'' behaviour and the property of a user and an item, respectively.
Unless there is a ``significant'' amount of interactions that cause a drift in a user's interest, for example from $C_2$ to $C_1$, ``unexpected" interactions should not have a strong influence on future behaviors~\cite{10.1145/1557019.1557072}. 
Rarely, a user might interact with items from distant contexts (e.g. $C_3$); for such a temporary case, a new instance can be projected without affecting other short-term embeddings.
This allows {\model} to reason based on embeddings that are closer to a query than exceptional cases.
Furthermore, {\model} gives the flexibility to use embedding histories as needed.
In addition, depending on the setting one can choose to discard old embeddings or store them in aggregated form.

The long-term embeddings, on the other hand, are updated and shifted to a new point, discarding the old ones. 
The dotted line in Fig.~\ref{fig:context_space_and_illustration} shows the trajectory of the long-term embedding of user $u$, starting from the inactive to active.
In a nutshell, these embeddings can be seen as aggregates of the short-term ones over time. 

In platforms like YouTube, LastFM, Spotify, the flexibility proposed by {\model} can be utilized to recommend multi-faceted sets of items, for example one based on short-term embeddings and another based on long-term ones.
This is in contrast to several studies that use a single embedding for recommendation.


\paragraph{Modeling long-term interactions} 
The overall preferences of users and the properties of items are captured by their long-term interactions.
To capture patterns in such interactions, we use identity-based embeddings of users and items that live in the same context space $\gS$.
That is, we use an embedding matrix $\mE \in \sR^{d \times U + I}$, which will be trained as interactions are observed; $\ve_u$ and $\ve_i$ denote the long-term embedding of user $u$ and item $i$.
To emphasize that $\ve_u$ and $\ve_i$ are conditioned on $\sL_u$ and $\sL_i$, we use the notation $\ve_u \vert \sL_u$ and $\ve_i \vert \sL_i$, respectively.



\paragraph{Modeling short-term interactions} Here the focus is in modeling recent interaction patterns of users and items that govern their follow-up actions.
To capture such patterns, we use short-term embeddings ${\vu(t)|\sL_u(t^<, k)}$ and ${\vi(t)|\sL_i(t^<, k)}$ for users and items, respectively, conditioned on their recent interactions.
In previous studies, short-term embeddings of users and items were recursively dependent on their respective previous short-term embeddings~\cite{kumar2019predicting,DBLP:journals/corr/DaiWTS16}.
The recursive nature of these algorithms inherently makes them expensive, as they would need to introduce a specialized algorithm for processing batches of interactions to avoid sequential processing.
For example, Kumar et al. had to introduce an algorithm called \textit{t-batch}, that process batches by respecting the temporal order~\cite{kumar2019predicting}.
Our design choice avoids such overhead by relying on the interaction histories rather than the previous short-term embeddings, which allows for ``simple'' batching.


\subsection{The proposed architecture}

\begin{figure}
    \centering
    \includegraphics[scale=0.32]{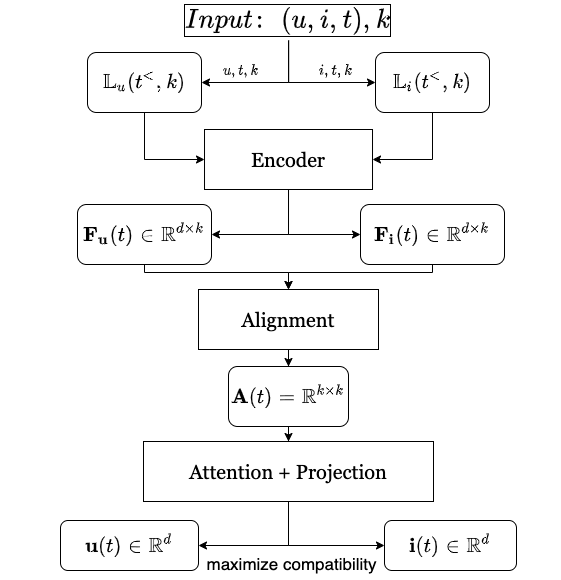}
    \caption{The architecture of \model: Parameters are updated at every event observation $(u, i, t)$. The update is carried out by inspecting the context (captured through long-term embeddings $\ve_{i_j} $ or $ \ve_{u_j}, j=1, ..., k$) of the $k$ recent events $\sL_u(t^<, k)$, $\sL_i(t^<, k)$ of $u$ and $i$ that happened just before $t$. The \emph{Encoder} generates high-level features that capture recurrence and temporal patterns within the $k$ events. The compatibility of each event in $\sL_u(t^<, k)$ and $\sL_i(t^<, k)$ is quantified using attention weights in the \emph{Attention} phase by leveraging a pair-wise \emph{Alignment} score. Finally, short-term embeddings $\vu(t)$ and $\vi(t)$ are obtained in the \emph{Projection} step as weighted sum of the high-level features}
    \label{fig:model_architecture}
\end{figure}

The complete architecture of {\model} is depicted in Fig.~\ref{fig:model_architecture}. 
The input of {\model} is given  by the observed interaction events and a hyper-parameter $k$ of the model.

\paragraph{Encoder} We process the user and item histories separately, using user and item encoders that share weights.
Again, this is in contrast to previous studies that use separate RNN modules that are dependent on previous short-term embeddings.
In \model, both the user and item encoder have the same structure;  for this reason, most of our discussion is related to the user encoder, while the item encoder is similar.
An encoder has two components:

\textit{The first component} of an encoder computes a signature embedding of the short-term history using the long-term embedding of the items (users) and the deltas as follows:
\begin{equation}\label{eq:user_signature}
    \mS_u(t) = f(\sL_u(t^<, k)) = [[\ve_{i_j}; \Delta_j] : (i_j, \Delta_j) \in \sL_u(t^<, k)]
\end{equation}
\begin{equation}\label{eq:item_signature}
    \mS_i(t) = f(\sL_i(t^<, k)) = [[\ve_{u_j};  \Delta_j] : (u_j, \Delta_j) \in \sL_i(t^<, k)]
\end{equation}
The simple, yet expressive and powerful trick used here is that  to compute the signature $\mS_u(t)$ at time $t$, Eq.(~\ref{eq:user_signature}) relies on the long-term embeddings of the $k$ most recent items that the user $u$ interacted with.
Equivalently, in Eq.~\ref{eq:item_signature}, the $k$ most recent users that interacted with the item $i$ are used to compute $\mS_i(t)$.
The key hypothesis is that the long-term or stationary embeddings of \emph{multiple items} is a strong signal for capturing a user's recent interest,
as each stationary embedding $\ve_{i_j} \in \mS_u(t)$ captures a sticking property or context (e.g. SciFi) of item $i_j$.
In addition, note that the signature at time $t$ contains information only from the \emph{past} (before time $t$), as we want to predict the present, time $t$.

Furthermore, it has been shown that the delay between interactions plays a significant role in predicting future interactions.
Thus, each long-term embedding is combined $[\cdot; \cdot]$ with $\Delta_j$ in the signature to increase the impact of fresh activities and decrease the importance of the stale ones.
Note that, some studies use a decay function for $\Delta_j$ instead, e.g. ${g(\Delta_j) = 1 / \log(e + \Delta_j)}$~\cite{ijcai2019-607,10.5555/3172077.3172393,10.1145/3159652.3159727}.
In our experiments we found no difference between these approaches, and hence we simply use $\Delta_j$.

\textit{The second component} of the encoder models recurring interaction and delay patterns in a history using shared and mutual RNN modules over the signatures, $\mS_u(t)$ and $\mS_i(t)$.
Empirically, Gated Recurrent units (GRU) tend to give better performance, thus we use GRU instead of the basic RNN.
Therefore, the standard GRU model for capturing recurrence in a signature $\mS(t)$ (user or item) slightly modified to integrate $\Delta_j$ is given as
\begin{align}
    \vz_j & = \sigma(\mW_{1z} \ve_j + \vb_{1z} + \mW_{2z} \Delta_j + \vb_{2z} + \mW_{3z} \vh_{j-1} + \vb_{3z} ) \\
    \vr_j & = \sigma(\mW_{1r} \ve_j + \vb_{1r} + \mW_{2r} \Delta_j + \vb_{2r} + \mW_{3r} \vh_{j-1} + \vb_{3r} ) \\
    \vn_j & = \tanh(\mW_{1n} \ve_j + \vb_{1n} + \mW_{2n} \Delta_j + \vb_{2n} + \vz_j \cdot (\mW_{3n} \vh_{j-1} + \vb_{3n})) \\
    \vh_j & = (1 - \vr_j) \cdot \vn_j + \vr_j \cdot \vh_{j-1}
\end{align}
where $\sigma$ is the sigmoid function and $\mW_{pq}$, $\vb_{pq}$, ${p \in \lbrace 1, 2, 3 \rbrace}$ and ${q \in \lbrace z, r, n \rbrace}$ are the parameters of the model shared by the encoders; $\ve_j$ corresponds to either $\ve_{i_j}$ or $\ve_{u_j}$ depending on the specified signature.
At each step $j$, a new hidden state $\vh_j$ is computed using the $j^{\textrm{th}}$ step inputs of $\mS(t)$, \emph{i.e.} the long-term embedding $\ve_j$ and $ \Delta_j$, and the previous hidden state $\vh_{j - 1}.$

Finally, we concatenate the hidden states of the GRU as
\begin{equation}
    \mF(t) = [\vh_1, \ldots, \vh_k]
\end{equation}
in order to obtain a high-level feature matrix of the signature at time $t$ that captures recurring interaction and delay patterns.
Again, depending on the encoder, $\mF(t)$ is either $\mF_u(t)$ or $\mF_i(t)$.


\paragraph{Alignment} Recall that both the user's and item's long-term embeddings live in the same space, and the high-level features $\mF_u(t)$ and $\mF_i(t)$ are derived based on such embeddings.
Thus, as shown in Eq.~\ref{eq:user_item_alignment}, the alignment component is used to inspect the compatibility between these features, to see how well the recent events of $u$ and $i$ agree contextually.
\begin{equation}\label{eq:user_item_alignment}
    \mA(t) = \tanh(\mF_u(t)^T \mF_i(t))
\end{equation}
We can interpret each row $j$ of $\mA(t) \in \sR^{k \times k}$ as a measure of context agreement between the $j^{\textrm{th}}$ item in the given user's $(u)$ short-term history with all the users in the given item's $(i)$ short-term history at time $t$.
In Eq.~\ref{eq:user_item_alignment}, similar to~\cite{santos2016attentive,kefato2020graph},  one can add more degree of freedom by introducing a trainable parameter $\Theta \in \sR^{d \times d}$ depending on the problem setting as in the following equation:
\begin{equation}\label{eq:user_item_alignment_more_df}
    \mA(t) = \tanh(\mF_u(t)^T \Theta \mF_i(t))
\end{equation}
However, we have empirically observed that for the problem at hand, fixing $\Theta$ to the identity matrix $\mI$ gives a better result.
When  Eq.~\ref{eq:user_item_alignment_more_df} is applied, {\model} tends to overfit faster even with a strong regularization; as a result, we opted for Eq.~\ref{eq:user_item_alignment} instead.
Hence, the only free parameters of {\model} are the long-term embedding $\mE$ and the GRU parameters.


\paragraph{Attention + Projection} 
Finally, in order to obtain embeddings that reflect short-term behaviours we pay attention to strong contextual agreements in $\mA(t)$, signaled by high scores, .
In other words, we want to investigate the compatibility between the recent interest of a user and the property of an item to understand where the agreement lies.
To this end, we compute attention weights for each item in the user's recent history (and vice-versa for each user in the item's recent history) using a column-wise ($\mX_{\bullet:}$) and row-wise ($\mX_{:\bullet}$) max-pooling as shown in Eq.~\ref{eq:colum_wise_pooling} and~\ref{eq:row_wise_pooling}, respectively.
\begin{equation}\label{eq:colum_wise_pooling}
    \Tilde{\vu}(t) = \max{\mA(t)_{\bullet:}}
\end{equation}
\begin{equation}\label{eq:row_wise_pooling}
    \Tilde{\vi}(t) = \max{\mA(t)_{:\bullet}}
\end{equation}
The $j^{\textrm{th}}$ component $\Tilde{\vu}_j(t)$ of the vector ${\tilde{\vu}(t) \in \sR^{k}}$ corresponds to the attention weight of the $j^{\textrm{th}}$ event, $(i_j, \Delta_j) \in \sL_u(t^<, k)$. 
It indicates:
\begin{itemize}
    \item[$\square$] the strongest alignment (contextual agreement) of the $j^{\textrm{th}}$ item $i_j$ from all the users in the short-term history $\sL_i(t^<, k)$ of the item $i$
    \item[$\square$] the hotness of the event
\end{itemize}
and it is the result of the column-wise pooling on the $j^{\textrm{th}}$ row, $\max(\mA(t)_{j:})$.
These two interpretations of the attention weights are based on the assumption that future activities are governed by recent actions and interest~\cite{ijcai2019-607,10.1145/3184558.3191526,10.5555/3172077.3172393}.
Inversely, stale events should have less impact on future interactions.

Equivalently, the $j^{\textrm{th}}$ component $\Tilde{\vi}_j(t)$ of ${\tilde{\vi}(t) \in \sR^{k}}$ represents the attention weights of the $j^{\textrm{th}}$ event,  $(u_j, \Delta_j)  \in \sL_i(t^<, k)$ and it is the result of the row-wise pooling on the $j^{\textrm{th}}$ column, $\max(\mA(t)_{:j})$.
The interpretation remains the same.

In this way, each item in the user history and each user in the item history are now scored in relation to their contextual agreement, from which we obtain the compatibility between the interacting user and item.
Alternatively, we have used mean-pooling in Eq.~\ref{eq:colum_wise_pooling} and~\ref{eq:row_wise_pooling} and empirically observed no difference.

At this point, we \emph{project} a new point representing the short-term interest and properties using the normalized attention weights.
Eq.~\ref{eq:user_projection} and~\ref{eq:item_projection} compute the user and item projection using the weighted sum of the features $\mF_u(t)$ and $\mF_i(t)$, respectively.
\begin{equation}\label{eq:user_projection}
    \vu(t) = \mF_u(t) \cdot \texttt{softmax}(\Tilde{\vu}(t)^T)
\end{equation}
\begin{equation}\label{eq:item_projection}
    \vi(t) = \mF_i(t) \cdot \texttt{softmax}(\Tilde{\vi}(t)^T)
\end{equation}
Both equations can be seen as feature selectors based on contextual agreement and freshness.
That is, they select those features that have a strong contextual agreement and are relatively new as indicated by the magnitude of the attention weights. 
The $\texttt{softmax}(\cdot)$ function gives us a distribution of weights for events in the short-term history of $u$ and $i$.
That is, fresh and contextually agreeing events will get weights close to 1, otherwise close to 0.
We argue that the model can learn in a way that weights are distributed in the aforementioned manner.
As desired, consequently, weighted features with weights close to 1 will govern the projections.
We consider $\vu(t)$ and $\vi(t)$ as predictions of the short-term embeddings of the user and item at time $t$, respectively.


\subsection{Training \model}
Similarly to previous work~\cite{kumar2019predicting}, {\model} predicts the user and item embeddings, albeit in a different manner.
Thus, we employ a similar loss function using mean squared error.
Our goal is to jointly train the long-term and short-term embeddings in order to bring the projection of frequently interacting items as close as possible.
To this end, we minimize the $L_2$ distance as 
\begin{equation}\label{eq:model_objective}
    \gL = \min \frac{1}{N} \sum_{(u, i, t) \in \sL_{\mathit{train}}}  \vert \vert \vu(t) - \vi(t) \vert \vert_2^2 + \gL_{reg}
\end{equation}
where $N$ is the batch size for batch training and $\sL_{\mathit{train}}$ is the observed event log in the training set.
The second term on the RHS of Eq.~\ref{eq:model_objective}, a regularization loss, is introduced to avoid the trivial solution of collapsing into a subspace.
It is motivated by the Laplacian eigenmaps method, which adds the constraint $\vu(t)^T \vi(t) = 1$ to avoid the collapse.
Therefore, we specify $\gL_{reg}$ as
\begin{equation}\label{eq:regularization}
    \gL_{reg} = \gamma \cdot \vert \vert \vv^T\vv - \mI \vert \vert_F^2
\end{equation}
where $\vv = [\vu(t); \vi(t)] \in \sR^{d \times 2}$ and $\gamma$ is a regularization coefficient.
$\gL_{reg}$ encourages points to be similar to themselves but not others.
Given that  we predict embeddings following~\cite{6789755,kumar2019predicting} as opposed to scores as in~\cite{DBLP:journals/corr/DaiWTS16}, we do not need for a contrastive loss in Eq.~\ref{eq:model_objective}.

Since our algorithm is designed in such a way that the short-term embeddings at time $t$ are not dependent on the ones at time $t - 1$, batching is straightforward and {\model} incurs in no overhead from batch processing unlike the work of Kumar et al.~\cite{kumar2019predicting}.
Together with design choices explained above, this makes {\model} efficient, as demonstrated in Section~\ref{sec:experiments}.


\subsection{{\model} for Static Networks}
{\model} requires only minor changes to be applicable to static interaction networks, as explained below.

The first obvious change is the lack of time, and consequently the lack of order; we consider $\sL$ to be an unordered set.
Thus, the notion of ``long-term'' and ``short-term'' interactions is meaningless.
Instead, the equivalent idea in static networks is ``global'' for ``long-term'' and ``context-aware'' for ``short-term''.
Global interactions are modeled as $(\ve_u \vert \sL_u$ or $\ve_i \vert \sL_i)$ using almost all the observed events in no specific order.
We refer to the corresponding embeddings as \emph{global embeddings}.
Similarly, context-aware interactions are modeled using \emph{context-aware embeddings} $\vu \vert \sL_u(k)$ or $\vi \vert \sL_i(k)$ conditioned on $k$ randomly sampled events.
The context-aware embeddings are in line with recent studies that argue against the adequacy of using a single embedding per node~\cite{Epasto_2019,Liu_2019,10.1145/3394486.3403293,kefato2020gossip,tu-etal-2017-cane}.
Each node, instead, is represented by multiple embeddings reflecting the multi-dimensional aspect of a node's interest or property.

%

Thus, the input is specified by each interaction $(u, i) \in \sL$ and $k$.
The user and item encoders take $\sL_u(k)$ and $\sL_i(k)$; encoding amounts to a simple embedding lookup and concatenation operation, to generate $\mF_u$ and $\mF_i$ ignoring the GRU model.
The followup steps are a straightforward application of the \emph{alignment} first, followed by \emph{attention + projection} to obtain the context-aware embeddings $\vu$ and $\vi$.

\section{Empirical Evaluation}
\label{sec:experiments}

We evaluate the performance of the proposed algorithm using three real-world temporal interaction networks and we compare {\model} against seven state-of-the-art baselines.

\begin{table*}[th!]
\begin{tabular}{|l|l|l|l|l|l|l||l|l|}
\hline
\multirow{2}{*}{\textbf{Method}} & \multicolumn{2}{c|}{\textbf{Reddit}} & \multicolumn{2}{c|}{\textbf{Wikipedia}} & \multicolumn{2}{c||}{\textbf{LastFM}} & \multicolumn{2}{l|}{\begin{tabular}[c]{@{}l@{}}\textbf{Minimum \% of improvement} \\ \textbf{of} {\model} \textbf{over method}\end{tabular}} \\ \cline{2-9} 
 & \textbf{MRR} & \textbf{Recall@10} & \textbf{MRR} & \textbf{Recall@10} & \textbf{MRR} & \textbf{Recall@10} & \textbf{MRR} & \textbf{Recall@10} \\ \hline
\lstm & 0.355 & 0.551 & 0.329 & 0.455 & 0.062 & 0.119 & 133.23 \% & 51.17 \% \\ \hline
\timelstm & 0.387 & 0.573 & 0.247 & 0.342 & 0.068 & 0.137 & 113.95 \% & 45.37 \% \\ \hline
\rrn & 0.603 & 0.747 & 0.522 & 0.617 & 0.089 & 0.182 & 37.13 \% & 11.51 \% \\ \hline
\latentcross & 0.421 & 0.588 & 0.424 & 0.481 & 0.148 & 0.227 & 96.67 \% & 41.67 \% \\ \hline
\ctdne & 0.165 & 0.257 & 0.035 & 0.056 & 0.01 & 0.01 & 401.81 \% & 224.12 \% \\ \hline
\deepcoevolve & 0.171 & 0.275 & 0.515 & 0.563 & 0.019 & 0.039 & 71.84 \% & 57.90 \% \\ \hline
\jodie & {\underline{0.726}} & \textbf{0.852} & {\underline{0.746}} & {\underline{0.822}} & {\underline{0.195}} & {\underline{0.307}} & 14.04 \% & -2.23 \% \\ \hline
\model & \textbf{0.828} & {\underline{0.833}} & \textbf{0.885} & \textbf{0.889} & \textbf{0.393} & \textbf{0.416} & - & - \\ \hline \hline
\textbf{\% gain over \jodie} & 14.04 \% & -2.23 \% & 18.63 \% & 8.15 \% & 101.53 \% & 35.50 \% & - & - \\ \hline
\end{tabular}
\caption{The comparison of the empirical results between {\model} and the baseline methods for the three temporal datasets. Bold and underline indicate best and second best performing algorithms, respectively.}
\label{tbl:next_item_prediction_result}
\end{table*}

\subsection{Datasets}

The three publicly available datasets we selected are the following:

\begin{itemize}
\item \textbf{Reddit}~\cite{kumar2019predicting}
contains post interactions by users on subreddits (items), over a period of one month.
The most active users (10,000) and items (1,000) are collected, with 672,447 interactions in total.
Actions are repeated 79\% of the time.

\item  \textbf{Wikipedia}~\cite{kumar2019predicting}
contains edit interactions by editors (users) on Wikipedia pages (items) over a period of one month.
8,227 editors with at least 5 edits and the 1,000 most edited pages  are included, for a total of 157,474 interactions.
Actions are repeated 61\% of the time.

\item  \textbf{LastFM}~\cite{kumar2019predicting}
contains listening activities by users on songs (items), over a period of one month, restricted to 1,000 users who listened to the 1,000 most-listened songs, with 1,293,103 interactions in total.
Actions are repeated  8.6\% of the time.
\end{itemize}

\subsection{Baselines}
We compare {\model} with seven state-of-the-art algorithms commonly used in recommender systems, grouped as follows:

\begin{itemize}
    \item \textbf{Sequence models} are different flavors of RNNs trained based on item-sequence data: \lstm, \timelstm~\cite{10.5555/3172077.3172393}, \rrn~\cite{10.1145/3018661.3018689}, \latentcross~\cite{10.1145/3159652.3159727}
    \item \textbf{Bipartite models} are baselines based on  bipartite interaction graph and employ mutually recursive RNNs: \deepcoevolve~\cite{dai2016deep}, \jodie~\cite{kumar2019predicting}.
    \item \textbf{Graph base model}: finally, we have \ctdne-based on contin\-uous-time graph embedding using temporal random walks.
\end{itemize}

\subsection{Next Item Prediction Experiment}

Based on observations of recent interactions with items, the goal is to predict the next item a user is likely to interact with.
This is what lies at the backbone of a number of RS.

\paragraph{Setting}
Similarly to Kumar et al.~\cite{kumar2019predicting}, data is partitioned by respecting the temporal order of events as training (80\%), validation (10\%), and test (10\%).
During training, we use the validation set to tune the model hyperparameters using Bayesian optimization.

During testing, given a ground-truth interaction $(u, i, t)$, {\model} predicts a ranked list of the top-$k$ items that $u$ will interact with at time $t$, based on previous interactions $\sL_u(t^<, k)$ and $\sL_i(t^<, k)$.
Since {\model} predicts short-term embeddings, as opposed to interaction probabilities, we can use an efficient nearest-neighbor search to predict the top-$k$ items.
We use mean reciprocal rank (MRR) and Recall@$k$ to measure the quality of the ranked list for the top-$k$ items, with $k=10$ and $k=1$.

\paragraph{Results}
Results are reported in Table~\ref{tbl:next_item_prediction_result}.
Since all the settings are exactly the same, the figures for all the baselines are directly taken from Kumar et al.~\cite{kumar2019predicting}.

\begin{table}[t!]
\begin{tabular}{|l|l|l|l|l|}
\hline
\multirow{2}{*}{\textbf{Method}} & \multicolumn{2}{c|}{\textbf{Reddit}} & \multicolumn{2}{c|}{\textbf{Wikipedia}} \\ \cline{2-5} 
 & \textbf{MRR} & \textbf{Recall@10} & \textbf{MRR} & \textbf{Recall@10}  \\ \hline
\jodie & 0.726 & 0.852 & 0.746 & 0.822  \\ \hline
\jodienf & 0.726 & 0.852 & 0.759 & 0.824  \\ \hline
\end{tabular}
\caption{{\jodie} vs \jodienf}
\label{tbl:jodie_vs_jodie_nf}
\vspace{-5mm}
\end{table}

{\model} outperforms all the baselines by a significant margin in all but one case.
Almost all the baselines have a huge gap between MRR and Recall@10, unlike the small gap of {\model}.
This shows that {\model} ranks the ground truth higher, while others simply detect it in lower positions in the top-10 predicted items.
For example, for the only case where {\jodie} beats {\model} by a small margin, we compare how {\jodie} and {\model} exactly match the ground truth, \emph{i.e.}, the Recall@1, and it is 0.648 for {\jodie} and 0.813 for {\model}.

We argue that \model's performance is mainly driven by the short-term embeddings projected based on the mutual attention mechanism (compatibility). 
As this enables {\model} to project a feature that is contextually related to the recent activities of the users and items, which is widely believed to govern future actions.

\paragraph{Effect of features}
One might ask, and rightly so, why not include a richer set of features  in \model, as in previous works~\cite{10.1145/3159652.3159727,kumar2019predicting,dai2016deep}.
First, some of these features (software client, page) are not easily accessible~\cite{10.1145/3159652.3159727}.
Other features, such as the textual content, could be easily integrated into our model without affecting the architecture; anyway, we found no difference for the three datasets.
To verify this, we have further investigated what happens when you remove textual features from the strongest baseline, {\jodie}.
As shown in Table~\ref{tbl:jodie_vs_jodie_nf}, {\jodienf} ({\jodie} with no features) performs as well as \jodie, if not better, for the two datasets with textual interaction features.

\begin{figure}[t!]
    \centering
    \includegraphics[scale=0.5]{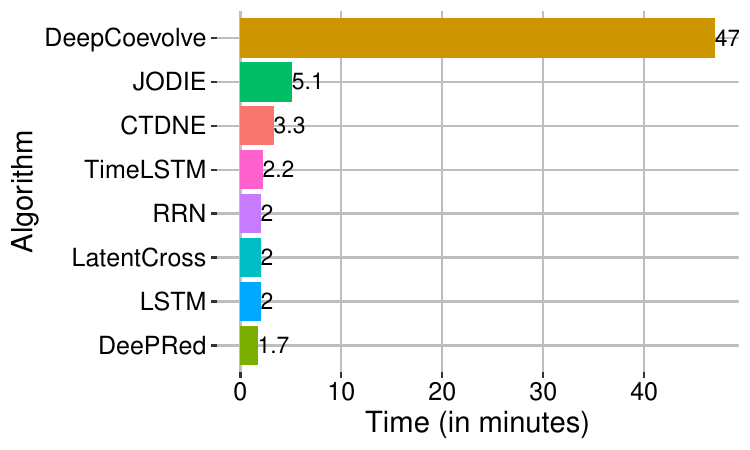}
    \caption{The computational time (in minutes) required to complete an epoch using the Reddit dataset.}
    \label{fig:run_time_analysis}
\end{figure}

\subsection{Runtime Experiment}

To empirically compare \model's efficiency, we measured the time needed to run the models.
In Fig.~\ref{fig:run_time_analysis}, we report the comparison between the methods for completing an epoch using the Reddit dataset.
We see that {\model} is much faster than all the baselines.
Since we are using the figures from~\cite{kumar2019predicting}, Fig.~\ref{fig:run_time_analysis} might not be a fair comparison as the machines are different.
Hence, we rerun {\jodie} on our machine and it took 15 minutes to complete the same epoch, showing that the speedup by {\model} is even better, more than an order of magnitude.

\subsection{Hyperparameter sensitivity experiment}
In this section, we analyze the effect of different hyperparameters of the methods on next item prediction. We simply compare {\model} with {\jodie}, since it is much better than all the other baselines.

\begin{figure}[ht!]
    \centering
    \includegraphics[scale=0.45]{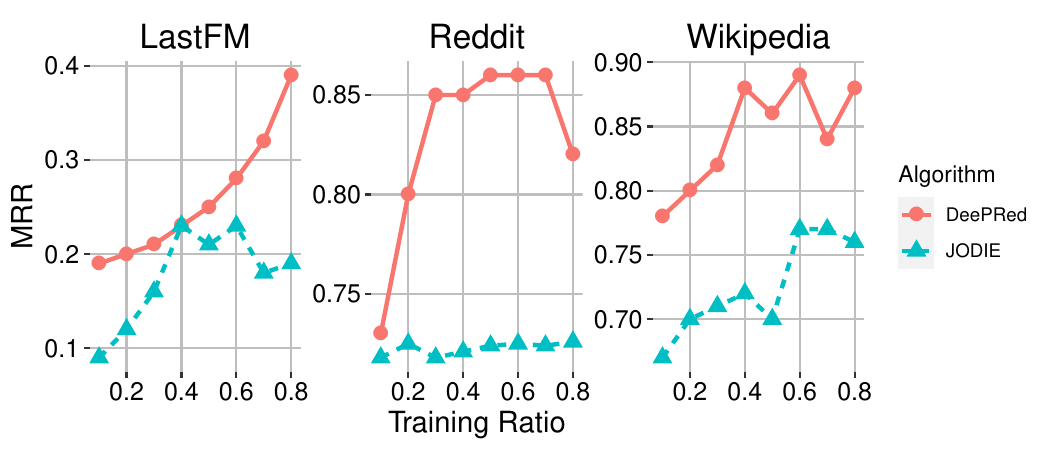}
    \caption{Effect of training proportion}
    \label{fig:training_ratio_analysis}
\end{figure}

\paragraph{Impact of training size} Despite their gap, as shown in Fig.~\ref{fig:training_ratio_analysis}, 60\% of the events are sufficient for both methods to effectively predict the next item on Reddit and Wikipedia. 
{\model} executed on LastFM, instead, keeps improving as repeated actions are sparse and patterns might emerge from observing more examples.

\begin{figure}[t!]
    \centering
    \includegraphics[scale=0.45]{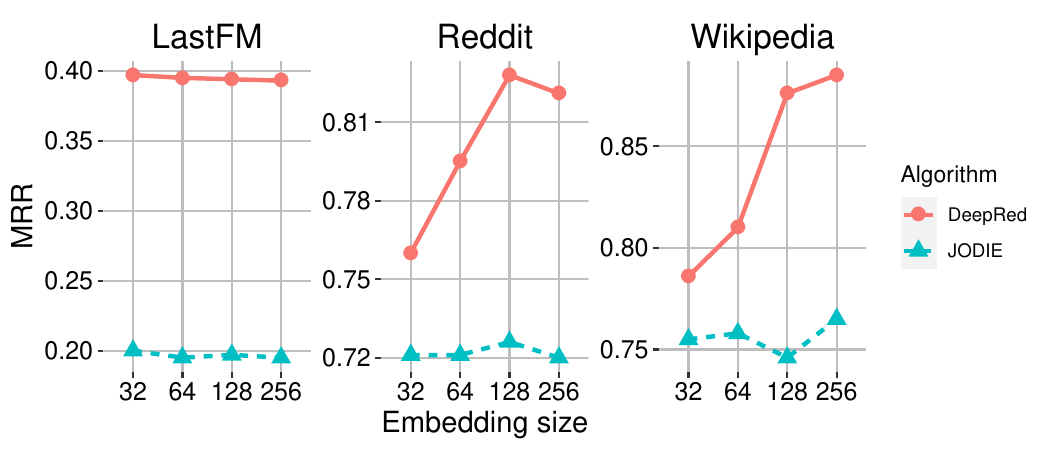}
    \caption{Effect of embedding size}
    \label{fig:embedding_size_analysis}
\end{figure}

\paragraph{Impact of embedding Size} 
Fig.~\ref{fig:embedding_size_analysis} shows the impact of the embedding size; for \model, 128 is an optimal value, while for {\jodie} this parameter has almost no influence.

\begin{figure}[t!]
    \centering
    \includegraphics[scale=0.45]{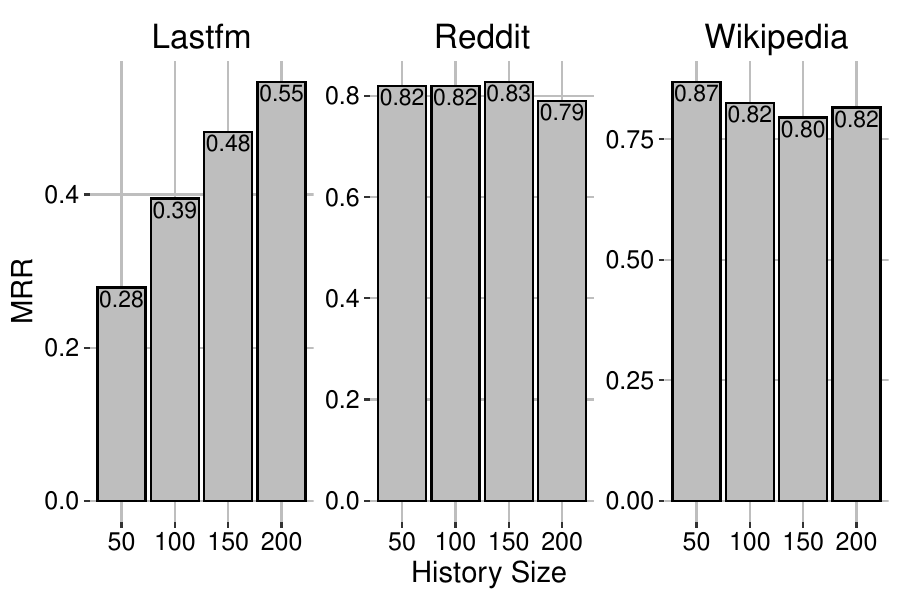}
    \caption{Effect of the short-term history size}
    \label{fig:history_size_analysis}
\end{figure}

\paragraph{Effect of $k$}
Parameter $k$, the 
number of short-term events in $\sL_u(t^<, k)$ and $\sL_i(t^<, k)$, affects {\model} only.
Our findings are reported in Fig.~\ref{fig:history_size_analysis}; we observe that $k$ has different effects across datasets.
In LastFM, increasing the number of events produces an improvement; in Reddit, there is no effect; in Wikipedia, a declining effect can be observed.
Recall that, actions are seldom repeated globally in LastFM, implying that repeated actions are locally sparse; for this reason, interaction patterns are detected by increasing the volume of retrospective observations.

\subsection{Static Networks' Experiment}

We discuss now our experiments carried out on  three static networks.
Although {\model} performs well, our goal here is to show its potential and flexibility, rather than report its superiority.

\paragraph{Datasets} We use the following static interaction networks:
\begin{itemize}
\item \textbf{MATADOR} (Manually Annotated Targets and Drugs Online Resource)~\cite{matador_dataset} is a drug-target interaction network, with 801 drugs (users) 2,901 targets (items), and 15,843 interactions.
\item \textbf{SIDER} (Side Effect Resource version 4.1)~\cite{sider_dataset} is a drug (user) and side-effects (item) association dataset. There are 639 users, 10,184 items and 174,977 interactions (associations).
\item \textbf{STEAM}~\cite{steam_game} is a popular PC gaming hub dataset, containing games (items) users have purchased. There are 12,393 users, 5,155 games, and 129,511 purchasing actions.
\end{itemize}

\paragraph{Baselines} We use four baselines grouped as follows:
\begin{itemize}
    \item \textbf{Context-aware}: \splitter~\cite{Epasto_2019} is a SOTA context-aware baseline; similarly to \model, it learns multiple embeddings of nodes for static networks.
    \item \textbf{Context-free} \deepwalk~\cite{DBLP:journals/corr/PerozziAS14}, \nodetovec~\cite{DBLP:journals/corr/GroverL16} and \lne~\cite{DBLP:journals/corr/TangQWZYM15} are popular baselines used for static network embedding 
\end{itemize}

\begin{figure}[th!]
    \centering
    \includegraphics[scale=0.40]{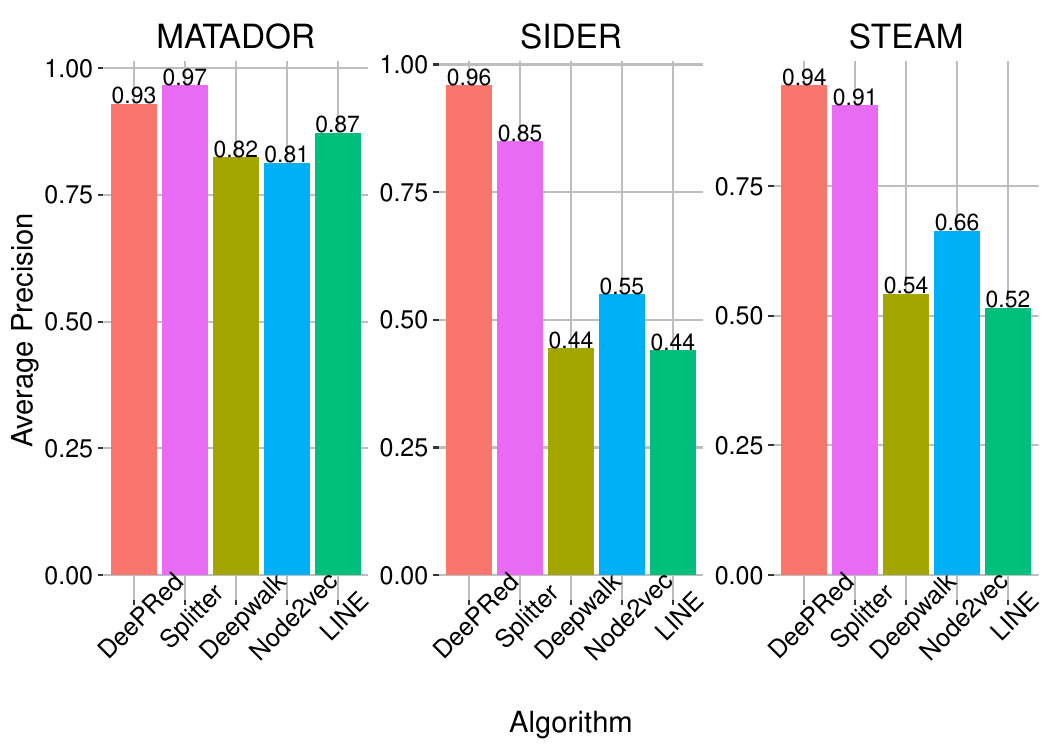}
    \caption{The Average Precision result for interaction prediction on static networks}
    \label{fig:interaction_prediction}
\end{figure}
The interaction prediction is executed as a link prediction task, where we create a random partition of the graph as training (60\%), validation (10\%), and test (30\%) sets.
In addition, we randomly sample non-existing (negative) interactions proportional to the test set (30\%).
An algorithm is trained on the training set and tuned on the validation set.
The average precision (AP), which summarizes the precision-recall curve is then computed based on a method's capacity to rank the test (true) over negative (false) interactions.
The results reported in Fig.~\ref{fig:interaction_prediction} shows that {\model} is comparable with a context-aware and much better than context-free baselines.

\section{Related Work}
\label{sec:related_work}
Factorization methods have significantly influenced the study of recommender systems (RS), more prominently since the Netflix prize competition.
However, as deep neural networks (DNNs) gained momentum across several domains, several studies have shown the effectiveness of DNNs in RS as well~\cite{10.1145/2959100.2959190,10.1145/3018661.3018719,10.1145/3331184.3331267,10.1145/3018661.3018689}. 
Early efforts used a vanilla DNN architecture by integrating crafted and learned features into the models~\cite{10.1145/2959100.2959190}.

As recurring patterns in user-item interactions are considered to be critical in recommending or predicting future activities, recurrent neural networks (RNNs) and its variants have been widely used in interaction prediction or RS.

\paragraph*{RNNs for Recommender Systems}
RNNs are inherently suited for modeling patterns in sequential data, such as language and time-series.
Due to their effectiveness, they have seen applicability in different areas, such as NLP, speech recognition, computer vision, and health--just to name a few.
Initial efforts in RS have employed RNNs by simply using a sequence of user actions in order to capture repeated user activity patterns, and model their preference or behavior~\cite{10.1145/3018661.3018689,10.1145/2988450.2988452,hidasi2015sessionbased}.
This approach has further been used to predict interesting items for users based on their preference, for example on platforms like YouTube, Spotify, LastFM.
However, standard RNNs and its variants (LSTM, GRU) can only capture recurrence and do not encode delay or interval between activities, which is an intrinsic nature of user behaviours.
This is because activities that are close to an event in time are more likely to trigger such event than the ones that are far apart.

\paragraph*{Time-based RNNs}
Motivated by the aforementioned need, extensions to RNNs (LSTM, GRU) have been introduced to account for time.
In addition to the existing gating mechanisms in RNNs, these studies have introduced different time-gating mechanisms to favor new events and discount the impact of old ones~\cite{10.5555/3172077.3172393,ijcai2019-607}.
Novelty or oldness refer to the delta in time, not to the position of events in a sequence.

\paragraph*{Mutual RNNs}
Closely related to our study, recently mutual RNNs for next item prediction have been proposed~\cite{kumar2019predicting,dai2016deep}.
A simple yet powerful aspect of these approaches is the bipartite temporal interaction network model, and the mutual RNN architecture that paved a way to examine user-item interaction dynamics.
However, besides the essential differences in modeling short-term embeddings of users and items,
{\model} is also different in using shared and non-recursive mutual RNNs.

\paragraph*{Other methods}
Besides RNNs, other methods such as graph neural networks (GNN) and transformers~ have also been employed in RS~\cite{10.5555/3295222.3295349}.
The former was introduced for neural collaborative-filtering and session-based RS~\cite{10.1145/3397271.3401142,Wu_2019,ijcai2019-547,10.1145/3331184.3331267}.
Due to the ever increasing impact of transformers for modeling sequential data, several studies proposed this model for predicting next basket items~\cite{kang2018selfattentive,10.1145/3357384.3357895,ijcai2019-547}.
Training transformers has proved to be much more efficient than RNNs, as they are highly parallelizable.
However, the core component of transformers--\emph{self-attention}--has the tendency to distribute attention weights and discounting impact from local dependencies~\cite{ijcai2019-547}.

\setcounter{table}{0}
\renewcommand{\thetable}{A\arabic{table}}

\begin{table*}[t!]
\begin{tabular}{|l|c|c|c|c|c|c|}
\hline
\multicolumn{1}{|c|}{\textbf{Parameter}} & \textbf{Reddit} & \textbf{Wikipedia} & \textbf{LastFM} & \textbf{Matador} & \textbf{Sider} & \textbf{Steam} \\ \hline
$k$ & 50 & 50 & 100 & 200 & 200 & 200 \\ \hline
$\gamma$ & 0.6 & 0.5 & 0.7 & 1.0 & 1.0 & 1.0 \\ \hline
Dropout rate & 0.6 & 0.7 & 0.8 & 0.3 & 0.3 & 0.3 \\ \hline
Learning rate & 0.0001 & 0.0001 & 0.0001 & 0.0001 & 0.0001 & 0.0001 \\ \hline
Embedding size & 128 & 128 & 128 & 128 & 128 & 128 \\ \hline
\end{tabular}

\caption{\model's hyperparameter configurations}
\label{lbl:model_configuration}
\end{table*}

\section{Conclusions and Future Work}
\label{sec:conclusion}

This study presents a novel algorithm called {\model} for next item prediction in temporal interaction networks.
Building up on recent achievements, {\model} captures the mutual interaction dynamics in the interactions between users and items.
We propose a simple yet powerful mechanism to model both user and item short-term preferences based on the their recent interaction history.
The history serves as proxy for the context of interaction in recent events.
We leverage the mechanism to avoid recursive dependency between consecutive short-term embeddings of a user or an item over time.
Our design enables {\model} to be effective in predicting next item interaction without compromising efficiency.

Our empirical finding on three real-world datasets demonstrate the effectiveness of {\model} over seven SOTA baselines by at least 14\% MRR measure.
In addition, {\model} is at least an order of magnitude faster than the best performing baselines.

We have also shown that the design of {\model} is flexible enough to accommodate static networks.
As a demonstration, we show how well it performs for interaction prediction over bio-chemical and gaming interaction networks.

Though maintaining multiple embeddings in {\model} is what lies behind its effectiveness, it comes at the cost of memory.
As GPU memory is expensive, this calls for an improved design for \model, that will be addressed in future work.

\appendix
\section{{\model} Configuration}
\label{apx:model_configuration}
Table~\ref{lbl:model_configuration} shows the final configurations of {\model} used to report the results in Section~\ref{sec:experiments}.
The experiments are executed on an NVIDIA QUADRO RTX 5000 GPU with NVLink, 3072 CUDA cores, and 16 GB GDDR6 memory.

\bibliographystyle{ACM-Reference-Format}
\bibliography{main}


\begin{thebibliography}{37}


\ifx \showCODEN    \undefined \def \showCODEN     #1{\unskip}     \fi
\ifx \showDOI      \undefined \def \showDOI       #1{#1}\fi
\ifx \showISBNx    \undefined \def \showISBNx     #1{\unskip}     \fi
\ifx \showISBNxiii \undefined \def \showISBNxiii  #1{\unskip}     \fi
\ifx \showISSN     \undefined \def \showISSN      #1{\unskip}     \fi
\ifx \showLCCN     \undefined \def \showLCCN      #1{\unskip}     \fi
\ifx \shownote     \undefined \def \shownote      #1{#1}          \fi
\ifx \showarticletitle \undefined \def \showarticletitle #1{#1}   \fi
\ifx \showURL      \undefined \def \showURL       {\relax}        \fi
\providecommand\bibfield[2]{#2}
\providecommand\bibinfo[2]{#2}
\providecommand\natexlab[1]{#1}
\providecommand\showeprint[2][]{arXiv:#2}

\bibitem[\protect\citeauthoryear{{Belkin} and {Niyogi}}{{Belkin} and
  {Niyogi}}{2003}]%
        {6789755}
\bibfield{author}{\bibinfo{person}{M. {Belkin}} {and} \bibinfo{person}{P.
  {Niyogi}}.} \bibinfo{year}{2003}\natexlab{}.
\newblock \bibinfo{title}{Laplacian Eigenmaps for Dimensionality Reduction and
  Data Representation}.
\newblock
\newblock


\bibitem[\protect\citeauthoryear{Beutel, Covington, Jain, Xu, Li, Gatto, and
  Chi}{Beutel et~al\mbox{.}}{2018}]%
        {10.1145/3159652.3159727}
\bibfield{author}{\bibinfo{person}{Alex Beutel}, \bibinfo{person}{Paul
  Covington}, \bibinfo{person}{Sagar Jain}, \bibinfo{person}{Can Xu},
  \bibinfo{person}{Jia Li}, \bibinfo{person}{Vince Gatto}, {and}
  \bibinfo{person}{Ed~H. Chi}.} \bibinfo{year}{2018}\natexlab{}.
\newblock \showarticletitle{Latent Cross: Making Use of Context in Recurrent
  Recommender Systems}. In \bibinfo{booktitle}{\emph{Proceedings of the
  Eleventh ACM International Conference on Web Search and Data Mining}} (Marina
  Del Rey, CA, USA) \emph{(\bibinfo{series}{WSDM '18})}.
  \bibinfo{publisher}{ACM}, \bibinfo{address}{New York, NY, USA},
  \bibinfo{pages}{46–54}.
\newblock


\bibitem[\protect\citeauthoryear{Buza and Peška}{Buza and Peška}{2017}]%
        {BUZA2017284}
\bibfield{author}{\bibinfo{person}{Krisztian Buza} {and}
  \bibinfo{person}{Ladislav Peška}.} \bibinfo{year}{2017}\natexlab{}.
\newblock \bibinfo{title}{Drug–target interaction prediction with Bipartite
  Local Models and hubness-aware regression}.
\newblock
\newblock


\bibitem[\protect\citeauthoryear{Covington, Adams, and Sargin}{Covington
  et~al\mbox{.}}{2016}]%
        {10.1145/2959100.2959190}
\bibfield{author}{\bibinfo{person}{Paul Covington}, \bibinfo{person}{Jay
  Adams}, {and} \bibinfo{person}{Emre Sargin}.}
  \bibinfo{year}{2016}\natexlab{}.
\newblock \showarticletitle{Deep Neural Networks for YouTube Recommendations}.
  In \bibinfo{booktitle}{\emph{Proceedings of the 10th ACM Conference on
  Recommender Systems}} (Boston, Massachusetts, USA)
  \emph{(\bibinfo{series}{RecSys '16})}. \bibinfo{publisher}{ACM},
  \bibinfo{address}{New York, NY, USA}, \bibinfo{pages}{191–198}.
\newblock
\showISBNx{9781450340359}


\bibitem[\protect\citeauthoryear{Dai, Wang, Trivedi, and Song}{Dai
  et~al\mbox{.}}{2016a}]%
        {dai2016deep}
\bibfield{author}{\bibinfo{person}{Hanjun Dai}, \bibinfo{person}{Yichen Wang},
  \bibinfo{person}{Rakshit Trivedi}, {and} \bibinfo{person}{Le Song}.}
  \bibinfo{year}{2016}\natexlab{a}.
\newblock \bibinfo{title}{Deep Coevolutionary Network: Embedding User and Item
  Features for Recommendation}.
\newblock
\newblock
\showeprint[arxiv]{1609.03675}~[cs.LG]


\bibitem[\protect\citeauthoryear{Dai, Wang, Trivedi, and Song}{Dai
  et~al\mbox{.}}{2016b}]%
        {DBLP:journals/corr/DaiWTS16}
\bibfield{author}{\bibinfo{person}{Hanjun Dai}, \bibinfo{person}{Yichen Wang},
  \bibinfo{person}{Rakshit Trivedi}, {and} \bibinfo{person}{Le Song}.}
  \bibinfo{year}{2016}\natexlab{b}.
\newblock \showarticletitle{Recurrent Coevolutionary Latent Feature Processes
  for Continuous-Time Recommendation}. In \bibinfo{booktitle}{\emph{Proceedings
  of the 1st Workshop on Deep Learning for Recommender Systems}} (Boston, MA,
  USA) \emph{(\bibinfo{series}{DLRS 2016})}. \bibinfo{publisher}{ACM},
  \bibinfo{address}{New York, NY, USA}, \bibinfo{pages}{29–34}.
\newblock
\showISBNx{9781450347952}


\bibitem[\protect\citeauthoryear{dos Santos, Tan, Xiang, and Zhou}{dos Santos
  et~al\mbox{.}}{2016}]%
        {santos2016attentive}
\bibfield{author}{\bibinfo{person}{Cicero dos Santos}, \bibinfo{person}{Ming
  Tan}, \bibinfo{person}{Bing Xiang}, {and} \bibinfo{person}{Bowen Zhou}.}
  \bibinfo{year}{2016}\natexlab{}.
\newblock \bibinfo{title}{Attentive Pooling Networks}.
\newblock
\newblock
\showeprint[arxiv]{1602.03609}~[cs.CL]


\bibitem[\protect\citeauthoryear{Epasto and Perozzi}{Epasto and
  Perozzi}{2019}]%
        {Epasto_2019}
\bibfield{author}{\bibinfo{person}{Alessandro Epasto} {and}
  \bibinfo{person}{Bryan Perozzi}.} \bibinfo{year}{2019}\natexlab{}.
\newblock \showarticletitle{Is a Single Embedding Enough? Learning Node
  Representations That Capture Multiple Social Contexts}. In
  \bibinfo{booktitle}{\emph{The World Wide Web Conference}} (San Francisco, CA,
  USA) \emph{(\bibinfo{series}{WWW '19})}. \bibinfo{publisher}{ACM},
  \bibinfo{address}{New York, NY, USA}, \bibinfo{pages}{394–404}.
\newblock


\bibitem[\protect\citeauthoryear{Grover and Leskovec}{Grover and
  Leskovec}{2016}]%
        {DBLP:journals/corr/GroverL16}
\bibfield{author}{\bibinfo{person}{Aditya Grover} {and} \bibinfo{person}{Jure
  Leskovec}.} \bibinfo{year}{2016}\natexlab{}.
\newblock \showarticletitle{Node2vec: Scalable Feature Learning for Networks}.
  In \bibinfo{booktitle}{\emph{Proceedings of the 22nd ACM SIGKDD International
  Conference on Knowledge Discovery and Data Mining}} (San Francisco,
  California, USA) \emph{(\bibinfo{series}{KDD '16})}.
  \bibinfo{publisher}{ACM}, \bibinfo{address}{New York, NY, USA},
  \bibinfo{pages}{855–864}.
\newblock
\showISBNx{9781450342322}


\bibitem[\protect\citeauthoryear{Hidasi, Karatzoglou, Baltrunas, and
  Tikk}{Hidasi et~al\mbox{.}}{2015}]%
        {hidasi2015sessionbased}
\bibfield{author}{\bibinfo{person}{Balázs Hidasi}, \bibinfo{person}{Alexandros
  Karatzoglou}, \bibinfo{person}{Linas Baltrunas}, {and}
  \bibinfo{person}{Domonkos Tikk}.} \bibinfo{year}{2015}\natexlab{}.
\newblock \bibinfo{title}{Session-based Recommendations with Recurrent Neural
  Networks}.
\newblock
\newblock


\bibitem[\protect\citeauthoryear{Jing and Smola}{Jing and Smola}{2017}]%
        {10.1145/3018661.3018719}
\bibfield{author}{\bibinfo{person}{How Jing} {and}
  \bibinfo{person}{Alexander~J. Smola}.} \bibinfo{year}{2017}\natexlab{}.
\newblock \showarticletitle{Neural Survival Recommender}. In
  \bibinfo{booktitle}{\emph{Proceedings of the Tenth ACM International
  Conference on Web Search and Data Mining}} (Cambridge, United Kingdom)
  \emph{(\bibinfo{series}{WSDM '17})}. \bibinfo{publisher}{Association for
  Computing Machinery}, \bibinfo{address}{New York, NY, USA},
  \bibinfo{pages}{515–524}.
\newblock


\bibitem[\protect\citeauthoryear{Kang and McAuley}{Kang and McAuley}{2018}]%
        {kang2018selfattentive}
\bibfield{author}{\bibinfo{person}{Wang-Cheng Kang} {and}
  \bibinfo{person}{Julian McAuley}.} \bibinfo{year}{2018}\natexlab{}.
\newblock \bibinfo{title}{Self-Attentive Sequential Recommendation}.
\newblock
\newblock


\bibitem[\protect\citeauthoryear{Kefato and Girdzijauskas}{Kefato and
  Girdzijauskas}{2020a}]%
        {kefato2020gossip}
\bibfield{author}{\bibinfo{person}{Zekarias~T. Kefato} {and}
  \bibinfo{person}{Sarunas Girdzijauskas}.} \bibinfo{year}{2020}\natexlab{a}.
\newblock \showarticletitle{Gossip and Attend: Context-Sensitive Graph
  Representation Learning}. In \bibinfo{booktitle}{\emph{Proceedings of the
  Fourteenth International {AAAI} Conference on Web and Social Media, {ICWSM}
  2020, Held Virtually, Original Venue: Atlanta, Georgia, USA, June 8-11,
  2020}}, \bibfield{editor}{\bibinfo{person}{Munmun~De Choudhury},
  \bibinfo{person}{Rumi Chunara}, \bibinfo{person}{Aron Culotta}, {and}
  \bibinfo{person}{Brooke~Foucault Welles}} (Eds.). \bibinfo{publisher}{{AAAI}
  Press}, \bibinfo{address}{Atlanta, Georgia, USA}, \bibinfo{pages}{351--359}.
\newblock
\urldef\tempurl%
\url{https://aaai.org/ojs/index.php/ICWSM/article/view/7305}
\showURL{%
\tempurl}


\bibitem[\protect\citeauthoryear{Kefato and Girdzijauskas}{Kefato and
  Girdzijauskas}{2020b}]%
        {kefato2020graph}
\bibfield{author}{\bibinfo{person}{Zekarias~T. Kefato} {and}
  \bibinfo{person}{Sarunas Girdzijauskas}.} \bibinfo{year}{2020}\natexlab{b}.
\newblock \bibinfo{title}{Graph Neighborhood Attentive Pooling}.
\newblock
\newblock
\showeprint[arxiv]{2001.10394}~[cs.LG]


\bibitem[\protect\citeauthoryear{Koren}{Koren}{2009}]%
        {10.1145/1557019.1557072}
\bibfield{author}{\bibinfo{person}{Yehuda Koren}.}
  \bibinfo{year}{2009}\natexlab{}.
\newblock \showarticletitle{Collaborative Filtering with Temporal Dynamics}. In
  \bibinfo{booktitle}{\emph{Proceedings of the 15th ACM SIGKDD International
  Conference on Knowledge Discovery and Data Mining}} (Paris, France)
  \emph{(\bibinfo{series}{KDD '09})}. \bibinfo{publisher}{Association for
  Computing Machinery}, \bibinfo{address}{New York, NY, USA},
  \bibinfo{pages}{447–456}.
\newblock


\bibitem[\protect\citeauthoryear{Kumar, Zhang, and Leskovec}{Kumar
  et~al\mbox{.}}{2019}]%
        {kumar2019predicting}
\bibfield{author}{\bibinfo{person}{Srijan Kumar}, \bibinfo{person}{Xikun
  Zhang}, {and} \bibinfo{person}{Jure Leskovec}.}
  \bibinfo{year}{2019}\natexlab{}.
\newblock \showarticletitle{Predicting Dynamic Embedding Trajectory in Temporal
  Interaction Networks}. In \bibinfo{booktitle}{\emph{Proceedings of the 25th
  ACM SIGKDD International Conference on Knowledge Discovery and Data Mining}}.
  \bibinfo{publisher}{Association for Computing Machinery},
  \bibinfo{address}{New York, NY, USA}, \bibinfo{pages}{1269–1278}.
\newblock
\showISBNx{9781450362016}


\bibitem[\protect\citeauthoryear{Liu, Tan, Li, Yang, Zhou, and Hu}{Liu
  et~al\mbox{.}}{2019}]%
        {Liu_2019}
\bibfield{author}{\bibinfo{person}{Ninghao Liu}, \bibinfo{person}{Qiaoyu Tan},
  \bibinfo{person}{Yuening Li}, \bibinfo{person}{Hongxia Yang},
  \bibinfo{person}{Jingren Zhou}, {and} \bibinfo{person}{Xia Hu}.}
  \bibinfo{year}{2019}\natexlab{}.
\newblock \bibinfo{booktitle}{\emph{Is a Single Vector Enough? Exploring Node
  Polysemy for Network Embedding}}.
\newblock \bibinfo{publisher}{Association for Computing Machinery},
  \bibinfo{address}{New York, NY, USA}, \bibinfo{pages}{932–940}.
\newblock
\showISBNx{9781450362016}
\urldef\tempurl%
\url{https://doi.org/10.1145/3292500.3330967}
\showURL{%
\tempurl}


\bibitem[\protect\citeauthoryear{M, I, LJ, and P}{M et~al\mbox{.}}{2015}]%
        {sider_dataset}
\bibfield{author}{\bibinfo{person}{Kuhn M}, \bibinfo{person}{Letunic I},
  \bibinfo{person}{Jensen LJ}, {and} \bibinfo{person}{Bork P}.}
  \bibinfo{year}{2015}\natexlab{}.
\newblock \bibinfo{title}{The SIDER database of drugs and side effects.}
\newblock
\newblock


\bibitem[\protect\citeauthoryear{Nguyen, Lee, Rossi, Ahmed, Koh, and
  Kim}{Nguyen et~al\mbox{.}}{2018}]%
        {10.1145/3184558.3191526}
\bibfield{author}{\bibinfo{person}{Giang~Hoang Nguyen},
  \bibinfo{person}{John~Boaz Lee}, \bibinfo{person}{Ryan~A. Rossi},
  \bibinfo{person}{Nesreen~K. Ahmed}, \bibinfo{person}{Eunyee Koh}, {and}
  \bibinfo{person}{Sungchul Kim}.} \bibinfo{year}{2018}\natexlab{}.
\newblock \showarticletitle{Continuous-Time Dynamic Network Embeddings}. In
  \bibinfo{booktitle}{\emph{Companion Proceedings of the The Web Conference
  2018}} (Lyon, France) \emph{(\bibinfo{series}{WWW '18})}.
  \bibinfo{publisher}{International World Wide Web Conferences Steering
  Committee}, \bibinfo{address}{Republic and Canton of Geneva, CHE},
  \bibinfo{pages}{969–976}.
\newblock
\showISBNx{9781450356404}
\urldef\tempurl%
\url{https://doi.org/10.1145/3184558.3191526}
\showDOI{\tempurl}


\bibitem[\protect\citeauthoryear{Perozzi, Al-Rfou, and Skiena}{Perozzi
  et~al\mbox{.}}{2014}]%
        {DBLP:journals/corr/PerozziAS14}
\bibfield{author}{\bibinfo{person}{Bryan Perozzi}, \bibinfo{person}{Rami
  Al-Rfou}, {and} \bibinfo{person}{Steven Skiena}.}
  \bibinfo{year}{2014}\natexlab{}.
\newblock \showarticletitle{DeepWalk: Online Learning of Social
  Representations}. In \bibinfo{booktitle}{\emph{Proceedings of the 20th ACM
  SIGKDD International Conference on Knowledge Discovery and Data Mining}} (New
  York, New York, USA) \emph{(\bibinfo{series}{KDD '14})}.
  \bibinfo{publisher}{Association for Computing Machinery},
  \bibinfo{address}{New York, NY, USA}, \bibinfo{pages}{701–710}.
\newblock
\showISBNx{9781450329569}


\bibitem[\protect\citeauthoryear{S, M, M, M, C, E, J, EG, A, LJ, R, R, RB, PE,
  P, and R.}{S et~al\mbox{.}}{2008}]%
        {matador_dataset}
\bibfield{author}{\bibinfo{person}{Günther S}, \bibinfo{person}{Kuhn M},
  \bibinfo{person}{Dunkel M}, \bibinfo{person}{Campillos M},
  \bibinfo{person}{Senger C}, \bibinfo{person}{Petsalaki E},
  \bibinfo{person}{Ahmed J}, \bibinfo{person}{Urdiales EG},
  \bibinfo{person}{Gewiess A}, \bibinfo{person}{Jensen LJ},
  \bibinfo{person}{Schneider R}, \bibinfo{person}{Skoblo R},
  \bibinfo{person}{Russell RB}, \bibinfo{person}{Bourne PE},
  \bibinfo{person}{Bork P}, {and} \bibinfo{person}{Preissner R.}}
  \bibinfo{year}{2008}\natexlab{}.
\newblock \bibinfo{title}{SuperTarget and Matador: resources for exploring
  drug-target relationships.}
\newblock
\newblock


\bibitem[\protect\citeauthoryear{Steam}{Steam}{0}]%
        {steam_game}
\bibfield{author}{\bibinfo{person}{Steam}.} \bibinfo{year}{0}\natexlab{}.
\newblock \bibinfo{title}{https://www.kaggle.com/tamber/steam-video-games}.
\newblock
\newblock


\bibitem[\protect\citeauthoryear{Sun, Liu, Wu, Pei, Lin, Ou, and Jiang}{Sun
  et~al\mbox{.}}{2019}]%
        {10.1145/3357384.3357895}
\bibfield{author}{\bibinfo{person}{Fei Sun}, \bibinfo{person}{Jun Liu},
  \bibinfo{person}{Jian Wu}, \bibinfo{person}{Changhua Pei},
  \bibinfo{person}{Xiao Lin}, \bibinfo{person}{Wenwu Ou}, {and}
  \bibinfo{person}{Peng Jiang}.} \bibinfo{year}{2019}\natexlab{}.
\newblock \showarticletitle{BERT4Rec: Sequential Recommendation with
  Bidirectional Encoder Representations from Transformer}. In
  \bibinfo{booktitle}{\emph{Proceedings of the 28th ACM International
  Conference on Information and Knowledge Management}} (Beijing, China)
  \emph{(\bibinfo{series}{CIKM '19})}. \bibinfo{publisher}{Association for
  Computing Machinery}, \bibinfo{address}{New York, NY, USA},
  \bibinfo{pages}{1441–1450}.
\newblock
\showISBNx{9781450369763}


\bibitem[\protect\citeauthoryear{Tan, Xu, and Liu}{Tan et~al\mbox{.}}{2016}]%
        {10.1145/2988450.2988452}
\bibfield{author}{\bibinfo{person}{Yong~Kiam Tan}, \bibinfo{person}{Xinxing
  Xu}, {and} \bibinfo{person}{Yong Liu}.} \bibinfo{year}{2016}\natexlab{}.
\newblock \showarticletitle{Improved Recurrent Neural Networks for
  Session-Based Recommendations}. In \bibinfo{booktitle}{\emph{Proceedings of
  the 1st Workshop on Deep Learning for Recommender Systems}} (Boston, MA, USA)
  \emph{(\bibinfo{series}{DLRS 2016})}. \bibinfo{publisher}{Association for
  Computing Machinery}, \bibinfo{address}{New York, NY, USA},
  \bibinfo{pages}{17–22}.
\newblock
\showISBNx{9781450347952}


\bibitem[\protect\citeauthoryear{Tang, Qu, Wang, Zhang, Yan, and Mei}{Tang
  et~al\mbox{.}}{2015}]%
        {DBLP:journals/corr/TangQWZYM15}
\bibfield{author}{\bibinfo{person}{Jian Tang}, \bibinfo{person}{Meng Qu},
  \bibinfo{person}{Mingzhe Wang}, \bibinfo{person}{Ming Zhang},
  \bibinfo{person}{Jun Yan}, {and} \bibinfo{person}{Qiaozhu Mei}.}
  \bibinfo{year}{2015}\natexlab{}.
\newblock \bibinfo{booktitle}{\emph{LINE: Large-Scale Information Network
  Embedding}}.
\newblock \bibinfo{publisher}{International World Wide Web Conferences Steering
  Committee}, \bibinfo{address}{Republic and Canton of Geneva, CHE},
  \bibinfo{pages}{1067–1077}.
\newblock
\showISBNx{9781450334693}


\bibitem[\protect\citeauthoryear{Tu, Liu, Liu, and Sun}{Tu
  et~al\mbox{.}}{2017}]%
        {tu-etal-2017-cane}
\bibfield{author}{\bibinfo{person}{Cunchao Tu}, \bibinfo{person}{Han Liu},
  \bibinfo{person}{Zhiyuan Liu}, {and} \bibinfo{person}{Maosong Sun}.}
  \bibinfo{year}{2017}\natexlab{}.
\newblock \showarticletitle{{CANE}: Context-Aware Network Embedding for
  Relation Modeling}. In \bibinfo{booktitle}{\emph{Proceedings of the 55th
  Annual Meeting of the Association for Computational Linguistics (Volume 1:
  Long Papers)}}. \bibinfo{publisher}{Association for Computational
  Linguistics}, \bibinfo{address}{Vancouver, Canada},
  \bibinfo{pages}{1722--1731}.
\newblock


\bibitem[\protect\citeauthoryear{Vaswani, Shazeer, Parmar, Uszkoreit, Jones,
  Gomez, Kaiser, and Polosukhin}{Vaswani et~al\mbox{.}}{2017}]%
        {10.5555/3295222.3295349}
\bibfield{author}{\bibinfo{person}{Ashish Vaswani}, \bibinfo{person}{Noam
  Shazeer}, \bibinfo{person}{Niki Parmar}, \bibinfo{person}{Jakob Uszkoreit},
  \bibinfo{person}{Llion Jones}, \bibinfo{person}{Aidan~N. Gomez},
  \bibinfo{person}{undefinedukasz Kaiser}, {and} \bibinfo{person}{Illia
  Polosukhin}.} \bibinfo{year}{2017}\natexlab{}.
\newblock \showarticletitle{Attention is All You Need}. In
  \bibinfo{booktitle}{\emph{Proceedings of the 31st International Conference on
  Neural Information Processing Systems}} (Long Beach, California, USA)
  \emph{(\bibinfo{series}{NIPS'17})}. \bibinfo{publisher}{Curran Associates
  Inc.}, \bibinfo{address}{Red Hook, NY, USA}, \bibinfo{pages}{6000–6010}.
\newblock
\showISBNx{9781510860964}


\bibitem[\protect\citeauthoryear{Wang, He, Wang, Feng, and Chua}{Wang
  et~al\mbox{.}}{2019}]%
        {10.1145/3331184.3331267}
\bibfield{author}{\bibinfo{person}{Xiang Wang}, \bibinfo{person}{Xiangnan He},
  \bibinfo{person}{Meng Wang}, \bibinfo{person}{Fuli Feng}, {and}
  \bibinfo{person}{Tat-Seng Chua}.} \bibinfo{year}{2019}\natexlab{}.
\newblock \showarticletitle{Neural Graph Collaborative Filtering}. In
  \bibinfo{booktitle}{\emph{Proceedings of the 42nd International ACM SIGIR
  Conference on Research and Development in Information Retrieval}} (Paris,
  France) \emph{(\bibinfo{series}{SIGIR'19})}. \bibinfo{publisher}{Association
  for Computing Machinery}, \bibinfo{address}{New York, NY, USA},
  \bibinfo{pages}{165–174}.
\newblock
\showISBNx{9781450361729}
\urldef\tempurl%
\url{https://doi.org/10.1145/3331184.3331267}
\showDOI{\tempurl}


\bibitem[\protect\citeauthoryear{Wang, Wei, Cong, Li, Mao, and Qiu}{Wang
  et~al\mbox{.}}{2020}]%
        {10.1145/3397271.3401142}
\bibfield{author}{\bibinfo{person}{Ziyang Wang}, \bibinfo{person}{Wei Wei},
  \bibinfo{person}{Gao Cong}, \bibinfo{person}{Xiao-Li Li},
  \bibinfo{person}{Xian-Ling Mao}, {and} \bibinfo{person}{Minghui Qiu}.}
  \bibinfo{year}{2020}\natexlab{}.
\newblock \showarticletitle{Global Context Enhanced Graph Neural Networks for
  Session-Based Recommendation}. In \bibinfo{booktitle}{\emph{Proceedings of
  the 43rd International ACM SIGIR Conference on Research and Development in
  Information Retrieval}} (Virtual Event, China) \emph{(\bibinfo{series}{SIGIR
  '20})}. \bibinfo{publisher}{Association for Computing Machinery},
  \bibinfo{address}{New York, NY, USA}, \bibinfo{pages}{169–178}.
\newblock
\showISBNx{9781450380164}
\urldef\tempurl%
\url{https://doi.org/10.1145/3397271.3401142}
\showDOI{\tempurl}


\bibitem[\protect\citeauthoryear{Wu, Ahmed, Beutel, Smola, and Jing}{Wu
  et~al\mbox{.}}{2017}]%
        {10.1145/3018661.3018689}
\bibfield{author}{\bibinfo{person}{Chao-Yuan Wu}, \bibinfo{person}{Amr Ahmed},
  \bibinfo{person}{Alex Beutel}, \bibinfo{person}{Alexander~J. Smola}, {and}
  \bibinfo{person}{How Jing}.} \bibinfo{year}{2017}\natexlab{}.
\newblock \showarticletitle{Recurrent Recommender Networks}. In
  \bibinfo{booktitle}{\emph{Proceedings of the Tenth ACM International
  Conference on Web Search and Data Mining}} (Cambridge, United Kingdom)
  \emph{(\bibinfo{series}{WSDM '17})}. \bibinfo{publisher}{Association for
  Computing Machinery}, \bibinfo{address}{New York, NY, USA},
  \bibinfo{pages}{495–503}.
\newblock
\showISBNx{9781450346757}


\bibitem[\protect\citeauthoryear{Wu, Tang, Zhu, Wang, Xie, and Tan}{Wu
  et~al\mbox{.}}{2019}]%
        {Wu_2019}
\bibfield{author}{\bibinfo{person}{Shu Wu}, \bibinfo{person}{Yuyuan Tang},
  \bibinfo{person}{Yanqiao Zhu}, \bibinfo{person}{Liang Wang},
  \bibinfo{person}{Xing Xie}, {and} \bibinfo{person}{Tieniu Tan}.}
  \bibinfo{year}{2019}\natexlab{}.
\newblock \showarticletitle{Session-Based Recommendation with Graph Neural
  Networks}.
\newblock \bibinfo{journal}{\emph{Proceedings of the AAAI Conference on
  Artificial Intelligence}}  \bibinfo{volume}{33} (\bibinfo{date}{Jul}
  \bibinfo{year}{2019}), \bibinfo{pages}{346–353}.
\newblock
\showISSN{2159-5399}
\urldef\tempurl%
\url{https://doi.org/10.1609/aaai.v33i01.3301346}
\showDOI{\tempurl}


\bibitem[\protect\citeauthoryear{Xu, Zhao, Liu, Sheng, Xu, Zhuang, Fang, and
  Zhou}{Xu et~al\mbox{.}}{2019}]%
        {ijcai2019-547}
\bibfield{author}{\bibinfo{person}{Chengfeng Xu}, \bibinfo{person}{Pengpeng
  Zhao}, \bibinfo{person}{Yanchi Liu}, \bibinfo{person}{Victor~S. Sheng},
  \bibinfo{person}{Jiajie Xu}, \bibinfo{person}{Fuzhen Zhuang},
  \bibinfo{person}{Junhua Fang}, {and} \bibinfo{person}{Xiaofang Zhou}.}
  \bibinfo{year}{2019}\natexlab{}.
\newblock \showarticletitle{Graph Contextualized Self-Attention Network for
  Session-based Recommendation}. In \bibinfo{booktitle}{\emph{Proceedings of
  the Twenty-Eighth International Joint Conference on Artificial Intelligence,
  {IJCAI-19}}}. \bibinfo{publisher}{International Joint Conferences on
  Artificial Intelligence Organization}, \bibinfo{address}{Macao, China},
  \bibinfo{pages}{3940--3946}.
\newblock
\urldef\tempurl%
\url{https://doi.org/10.24963/ijcai.2019/547}
\showDOI{\tempurl}


\bibitem[\protect\citeauthoryear{Yang, Pal, Zhai, Pancha, Han, Rosenberg, and
  Leskovec}{Yang et~al\mbox{.}}{2020}]%
        {10.1145/3394486.3403293}
\bibfield{author}{\bibinfo{person}{Carl Yang}, \bibinfo{person}{Aditya Pal},
  \bibinfo{person}{Andrew Zhai}, \bibinfo{person}{Nikil Pancha},
  \bibinfo{person}{Jiawei Han}, \bibinfo{person}{Charles Rosenberg}, {and}
  \bibinfo{person}{Jure Leskovec}.} \bibinfo{year}{2020}\natexlab{}.
\newblock \showarticletitle{MultiSage: Empowering GCN with Contextualized
  Multi-Embeddings on Web-Scale Multipartite Networks}. In
  \bibinfo{booktitle}{\emph{Proceedings of the 26th ACM SIGKDD International
  Conference on Knowledge Discovery and Data Mining}}
  \emph{(\bibinfo{series}{KDD '20})}. \bibinfo{publisher}{ACM},
  \bibinfo{address}{New York, NY, USA}, \bibinfo{pages}{2434–2443}.
\newblock
\showISBNx{9781450379984}
\urldef\tempurl%
\url{https://doi.org/10.1145/3394486.3403293}
\showDOI{\tempurl}


\bibitem[\protect\citeauthoryear{You, McLeod, and Hu}{You
  et~al\mbox{.}}{2019}]%
        {YOU201990}
\bibfield{author}{\bibinfo{person}{Jiaying You}, \bibinfo{person}{Robert~D.
  McLeod}, {and} \bibinfo{person}{Pingzhao Hu}.}
  \bibinfo{year}{2019}\natexlab{}.
\newblock \bibinfo{title}{Predicting drug-target interaction network using deep
  learning model}.
\newblock
\newblock


\bibitem[\protect\citeauthoryear{Zhang, Yang, Ivy, and Chi}{Zhang
  et~al\mbox{.}}{2019}]%
        {ijcai2019-607}
\bibfield{author}{\bibinfo{person}{Yuan Zhang}, \bibinfo{person}{Xi Yang},
  \bibinfo{person}{Julie Ivy}, {and} \bibinfo{person}{Min Chi}.}
  \bibinfo{year}{2019}\natexlab{}.
\newblock \showarticletitle{ATTAIN: Attention-based Time-Aware LSTM Networks
  for Disease Progression Modeling}. In \bibinfo{booktitle}{\emph{Proceedings
  of the Twenty-Eighth International Joint Conference on Artificial
  Intelligence, {IJCAI-19}}}. \bibinfo{publisher}{International Joint
  Conferences on Artificial Intelligence Organization},
  \bibinfo{address}{Macao, China}, \bibinfo{pages}{4369--4375}.
\newblock
\urldef\tempurl%
\url{https://doi.org/10.24963/ijcai.2019/607}
\showDOI{\tempurl}


\bibitem[\protect\citeauthoryear{Zhu, Li, Liao, Wang, Guan, Liu, and Cai}{Zhu
  et~al\mbox{.}}{2017}]%
        {10.5555/3172077.3172393}
\bibfield{author}{\bibinfo{person}{Yu Zhu}, \bibinfo{person}{Hao Li},
  \bibinfo{person}{Yikang Liao}, \bibinfo{person}{Beidou Wang},
  \bibinfo{person}{Ziyu Guan}, \bibinfo{person}{Haifeng Liu}, {and}
  \bibinfo{person}{Deng Cai}.} \bibinfo{year}{2017}\natexlab{}.
\newblock \showarticletitle{What to Do next: Modeling User Behaviors by
  Time-LSTM}. In \bibinfo{booktitle}{\emph{Proceedings of the 26th
  International Joint Conference on Artificial Intelligence}}
  \emph{(\bibinfo{series}{IJCAI’17})}. \bibinfo{publisher}{AAAI Press},
  \bibinfo{address}{Melbourne, Australia}, \bibinfo{pages}{3602–3608}.
\newblock
\showISBNx{9780999241103}


\bibitem[\protect\citeauthoryear{Zitnik, Agrawal, and Leskovec}{Zitnik
  et~al\mbox{.}}{2018}]%
        {10.1093/bioinformatics/bty294}
\bibfield{author}{\bibinfo{person}{Marinka Zitnik}, \bibinfo{person}{Monica
  Agrawal}, {and} \bibinfo{person}{Jure Leskovec}.}
  \bibinfo{year}{2018}\natexlab{}.
\newblock \bibinfo{title}{{Modeling polypharmacy side effects with graph
  convolutional networks}}.
\newblock
\newblock
\urldef\tempurl%
\url{https://doi.org/10.1093/bioinformatics/bty294}
\showDOI{\tempurl}


\end{thebibliography}

\end{document}